\documentclass[letterpaper]{article} 
\usepackage{aaai24}  
\usepackage{times}  
\usepackage{helvet}  
\usepackage{courier}  
\usepackage[hyphens]{url}  
\usepackage{graphicx} 
\usepackage{natbib}  
\usepackage{caption} 
\usepackage{algorithm}
\usepackage{algorithmic}
\usepackage{amssymb}
\usepackage{amsmath,amsfonts}

\usepackage{tabularx}

\usepackage{newfloat}
\usepackage{listings}
\DeclareCaptionStyle{ruled}{labelfont=normalfont,labelsep=colon,strut=off} 
\floatstyle{ruled}
\newfloat{listing}{tb}{lst}{}
\floatname{listing}{Listing}
\title{Autoregressive Omni-Aware Outpainting for Open-Vocabulary \\ 360-Degree Image Generation}
\author {
    Zhuqiang Lu\textsuperscript{\rm 1},
    Kun Hu\textsuperscript{\rm 1,}\thanks{Corresponding author.}, 
    Chaoyue Wang\textsuperscript{\rm 2}, 
    Lei Bai\textsuperscript{\rm 3}, 
    Zhiyong Wang\textsuperscript{\rm 1}
}
\affiliations {
    \textsuperscript{\rm 1}The University of Sydney\\
    \textsuperscript{\rm 2}JD.com\\
    \textsuperscript{\rm 3}Shanghai AI Laboratory\\
    zhuqiang.lu@sydney.edu.au, kun.hu@sydney.edu.au, chaoyue.wang@outlook.com,
    baisanshi@gmail.com, zhiyong.wang@sydney.edu.au
}

\usepackage{bibentry}

\begin{document}

\maketitle

\begin{abstract}
A 360-degree (omni-directional) image provides an all-encompassing spherical view of a scene. Recently, there has been an increasing interest in synthesising 360-degree images from conventional narrow field of view (NFoV) images captured by digital cameras and smartphones, for providing immersive experiences in various scenarios such as virtual reality. Yet, existing methods typically fall short in synthesizing intricate visual details or ensure the generated images align consistently with user-provided prompts. In this study, autoregressive omni-aware generative network (AOG-Net) is proposed for 360-degree image generation by outpainting an incomplete 360-degree image progressively with NFoV and text guidances joinly or individually. This autoregressive scheme not only allows for deriving finer-grained and text-consistent patterns by dynamically generating and adjusting the process but also offers users greater flexibility to edit their conditions throughout the generation process. A global-local conditioning mechanism is devised to comprehensively formulate the outpainting guidance in each autoregressive step. Text guidances, omni-visual cues, NFoV inputs and omni-geometry are encoded and further formulated with cross-attention based transformers into a global stream and a local stream into a conditioned generative backbone model. As AOG-Net is compatible to leverage large-scale models for the conditional encoder and the generative prior, it enables the generation to use extensive open-vocabulary text guidances. Comprehensive experiments on two commonly used 360-degree image datasets for both indoor and outdoor settings demonstrate the state-of-the-art performance of our proposed method. Our code is available at \url{https://github.com/zhuqiangLu/AOG-NET-360}.

\end{abstract}

\section{Introduction}

A 360-degree (omni-directional) image offers a comprehensive spherical view of a scene and provides users the freedom to explore any direction from a singular view point. They have revolutionized the way that users consume, interact with, and produce visual content. 
Yet, the exclusive reliance on specialized cameras to capture these images poses significant challenges for their widespread adoption, limiting the scalability and accessibility of creating immersive content for broader audiences. 
In contrast, given the vast quantity of Narrow Field of View (NFoV) images captured daily via mobile phones and digital cameras, there has been a growing interest in transforming these conventional images into 360-degree panoramic visuals. By converting these NFoV images, extensive visual databases can be leveraged and enable more immersive experiences for the applications in Virtual Reality and Augmented Reality across various domains such as tourism, entertainment and education.

\begin{figure}
\centering
\includegraphics[width=1 \columnwidth]{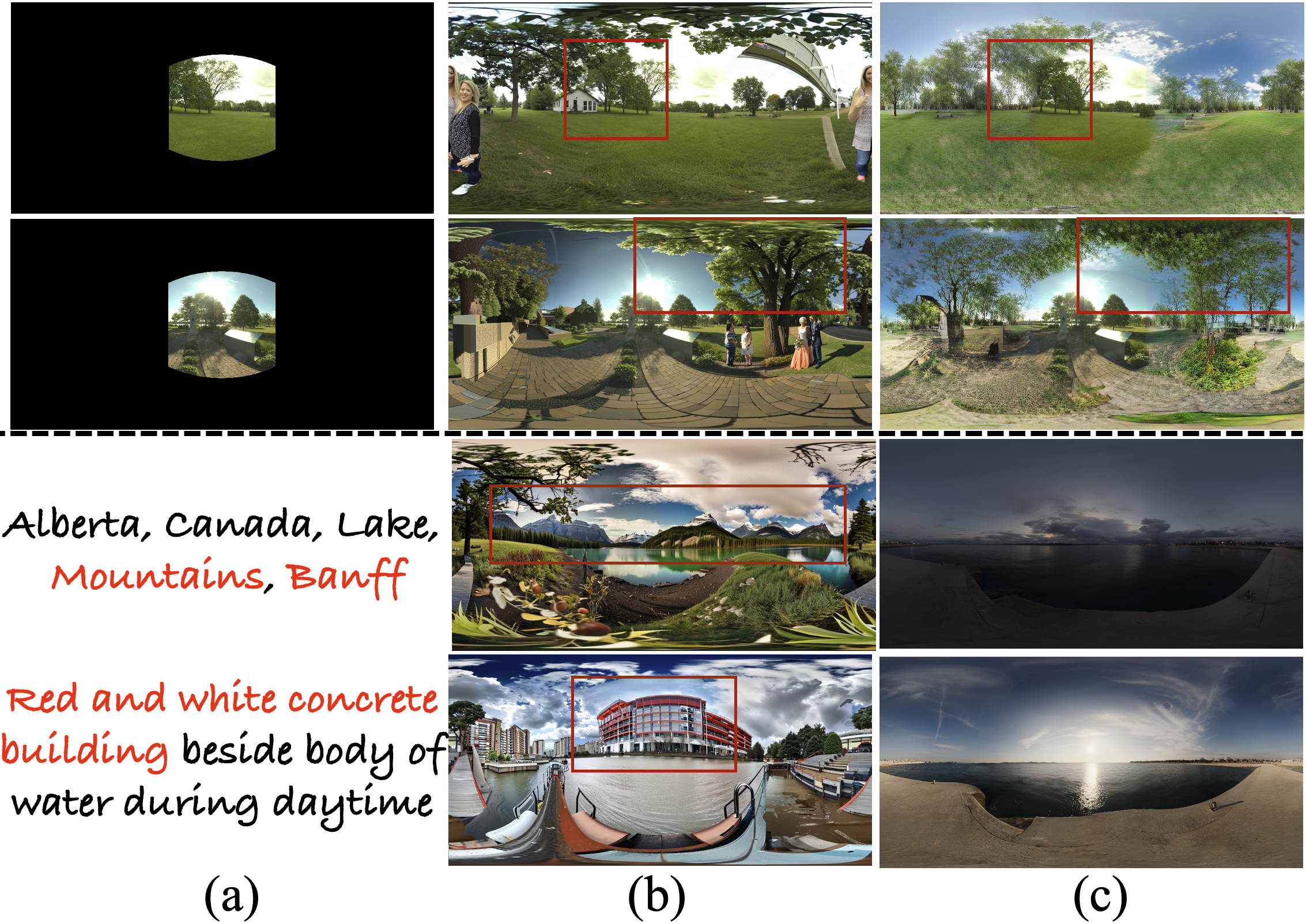} 
\caption{Examples of 360-degree image generation, showcasing the limitation of existing methods compared to ours. The top part above the dashed line depicts an NFoV-guided example and the bottom part below the dashed line is for a text-guided example. (a) Input condition. (b) Ours (AOG-Net). (c) Top - OmniDreamer \cite{akimoto2022diverse} and Bottom - Text2Light\cite{chen2023text2light}. }
\label{fig:demo}
\end{figure}

In recent years,  deep learning methods have been explored to generate photo-realistic 360-degree images. 
For instance, OmniDreamer \cite{akimoto2022diverse} formulates a 360-degree image generation pipeline with a VQGAN \cite{esser2020taming} by treating NFoV images as incomplete 360-degree images.
Conditioned on text guidances, Text2Light \cite{chen2023text2light} introduces two VQGANs for a global-to-local modelling strategy in pursuit of generating high-resolution 360-degree images. 
ImmerseGAN \cite{Dastjerdi_2022} applies domain adaptation methods on pretrained GANs, which can be conditioned on both NFoV images and text guidances. 
While these methods show encouraging performance, the challenge remains regarding the usage of given NFoV images and user-provided open vocabulary text guidances individually or jointly for enhanced control in 360-degree image generation. 
Specifically, the existing methods typically fall short in synthesizing intricate visual details as shown in the top part of Fig. \ref{fig:demo}, where the details are vague or missing with OmniDreamer compared to our approach, which are highlighted in the red bounding boxes. 
Moreover, the generated images and user-provided text guidances tend to be inconsistent, especially under an open-vocabulary setting, as depicted in the bottom part of Fig. \ref{fig:demo} by comparing Text2Light and our solution.

In this study, we propose a novel autoregressive omni-aware generative network (AOG-Net) for generating 360-degree images conditioned on open vocabulary text guidances and given NFoV images jointly or individually. 
Overall, the generation is formulated as an autoregressive stochastic process to outpaint an incomplete 360-degree image progressively, in which each step outpaints a local region under its corresponding NFoV view. 
This autoregressive scheme not only allows for deriving finer-grained and prompt-consistent patterns by dynamically observing and adjusting the generation process but also offers users greater flexibility to modify or introduce new conditions throughout the generation process. Furthermore, a global-local conditioning mechanism is devised to comprehensively formulate the outpainting guidance for each autoregression step. 
Text prompts, omni-visual cues, NFoV inputs and omni-geometry are encoded and further formulated with cross-attention based transformers into a global stream and a local stream for a conditioned generative backbone model. This study further explores the potential to leverage large-scale models for the conditional encoder and the generative prior, which helps complete the generation using open-vocabulary prompts. Comprehensive experiments on two commonly used 360-degree image datasets for both indoor and outdoor settings demonstrate the state-of-the-art performance of our proposed method. 

In summary, the key contributions of this study are:
\begin{itemize}
\item A novel autoregressive outpainting approach is proposed to produce photo-realistic 360-degree images by dynamically adjusting the generation process for improved finer-grained details and prompt-consistency. 
\item A global-local conditioning mechanism is devised to formulate the guidance encompassing open-vocab text guidances, omni-visual cues, NFoV inputs and omni-geometry with cross-attention based transformers. 
\item Comprehensive experiments were conducted on two commonly used benchmarks, demonstrating the state-of-the-art performance of AOG-Net in both indoor and outdoor settings with as few as 40 training samples. 
\end{itemize}

\begin{figure*}[t]
\centering
\includegraphics[width=1.\textwidth]{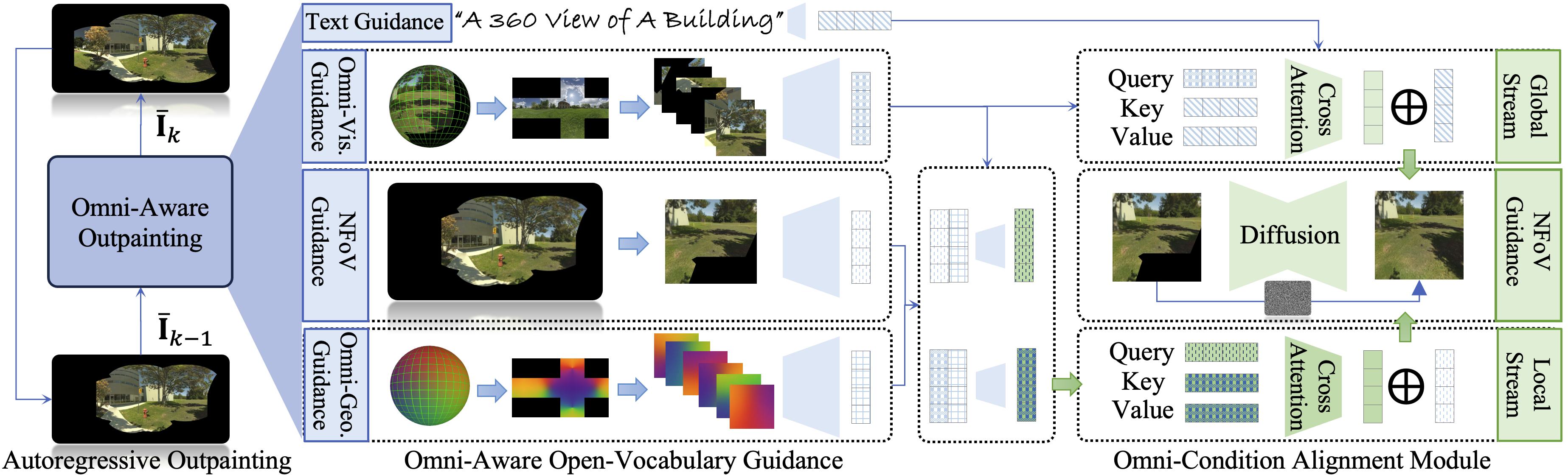} 
\caption{Illustration of the proposed AOG-Net architecture.}
\label{fig:overview}
\end{figure*}

\section{Related Work}
\label{sec:related}

We first review the studies in both the field of 360-degree image generation and the field of image outpainting which are relevant to our study. As our work takes image and text guidances as conditions, we further review the related studies on conditional image generation.

\subsection{360-Degree Image Generation}
Unlike general NFoV images, 360-degree image generation requires to take the omni-directional continuity into account. 
Early studies, for example, \cite{sengupta2019neural} estimates a coarse 360-degree image from an NFoV image with inverse rendering technique, which ignores such geometrical continuity and generates 360-degree images lack of fine details. 
To address this, 360IC \cite{360ic} and SIG-SS \cite{hara2020spherical} were proposed to improve geometrical continuity by taking the intrinsic horizontal cyclicity into consideration and encoding it as positional conditions to connect the two ends of 360-degree images in equirectangular representations.
EnvMapNet \cite{somanath2021hdr} improves visual quality of the outpainted 360-degree images by introducing a projection loss and a clustering loss for accurate lighting and shadowing. 
OmniDreamer \cite{akimoto2022diverse} was further developed by leveraging the Taming-Transformer \cite{esser2020taming}, where a circular inference scheme was introduced to fit the intrinsic horizontal cyclicity for 360-degree image synthesis, conditioned on provided NFoV images, yielding diverse and photo-realistic results. 
However, OmniDreamer is limited to a single condition where only an initial NFoV image is accepted, while the controllability of the overall synthesis process is limited. 
ImmenseGAN \cite{Dastjerdi_2022} aims for finer controllability over the outpainting by introducing a text guidances to fine-tune a generative model with a large-scale private text-image pair dataset. 
Due to the lack of public text-image paired dataset, Text2Light \cite{chen2023text2light} introduces a zero-shot text-guided 360-degree image synthesis pipeline without using initial NFoV images, in which a pre-trained CLIP model is adopted \cite{radford2021learning}.

However, the existing methods typically fall short in synthesizing intricate visual details and inconsistencies can be observed between generated images and user-provided text guidances, especially under an open-vocabulary setting, which demands further mechanisms to address these issues. 

\subsection{Image Outpainting}
Image outpainting, a fundamental task in computer vision, focuses on expanding the unknown regions outside the primary known content. Unlike inpainting, outpainting may not be able to leverage information from pixels adjacent to the unknown area \cite{sabini2018painting, vanhoorick2020image, wang2019srn}, as seen in inpainting methods. 
In \cite{wang2019srn}, the semantic information of incomplete images was utilized to guide a GAN for outpainting. 
In \cite{yao2022qotr}, a query-based outpainting method was proposed, where an image is divided into small patches and the patches with unknown pixels are completed by taking the conditions from both distant and neighbour patches into account. 
In an iterative manner, \cite{gardias2020enhanced} extends one side of the a regular image for outpainting step by step, using the context of the past generation as guidance. 
\cite{lu2021bridging} delves into the idea of synthesizing unknown regions by exploiting the correlations between distant image patches to establish the global semantics of known pixels.
Similarly, in \cite{esser2020taming, chang2022maskgit}, image outpainting methods were studied with transformers \cite{vaswani2023attention}, which predict the most probable pixel value recursively. However, these conventional outpainting methods do not account for the unique omni-directional continuity inherent to 360-degree images, often leading to discontinuities and artifacts.

\subsection{Conditional Image Generation}
Conditional image generation refers to the synthesis of images based on specific conditions, such as text prompts \cite{rombach2022highresolution, kang2023scaling}, semantic maps \cite{esser2020taming, chang2022maskgit} and audio cues \cite{yariv2023audiotoken}. 
For instance, \cite{isola2018imagetoimage} achieves conditional image generation using a conditional GAN \cite{mirza2014conditional} to formulate the joint probability of images and conditions. 
\cite{imaggpt, esser2020taming} treat an image as a sequence of pixels and therefore generate pixels in an iterative manner.  
Building on the success of diffusion methods in image generation \cite{ho2020denoising, song2022denoising}, various conditional diffusion models have been investigated. For example, \cite{dhariwal2021diffusion} introduces  an auxiliary classifier to guide the generation of images within a specific category. \cite{ho2022classifierfree} presents a unified framework for conditional generation using diffusion models, introducing a mechanism to control the correlation between the generated image and its input guidance.
However, alignment of these conditions with omni-directional geometry is not trivial and further omni-aware alignment strategy is required.

\section{Methodology}
As shown in Fig. \ref{fig:overview}, our proposed AOG-Net for 360-degree image generation follows an autoregressive manner by outpainting a local region progressively. In each step, a global-local conditioning mechanism is introduced to formulate text, omni-visual, NFoV and omni-geometry guidances with cross-attention based transformers into a global stream and a local stream. Such conditions are further adopted a backbone generative prior for the outpainting. 
The details of these components are discussed in this section.

\subsection{360-Degree Images \& Problem Formulation} 
\begin{figure}[t]
\centering
\includegraphics[width=1 \columnwidth]{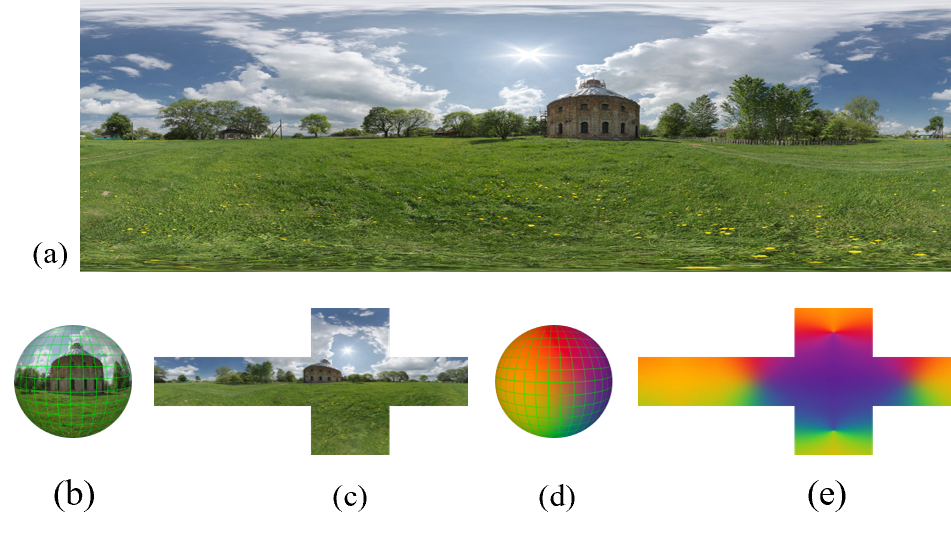} 
\caption{Different representations of a 360-degree image. (a) Equirectangular projection. (b) Spherical representation. (c) Cubemap projection. (d) A spherical representation with geometry coordinates. (e) Geometry projection on cubemap. }
\label{fig:representation}
\end{figure}

Given a 360-degree image, denoted as $\mathbf{I}$, there are three typical representations as shown in Fig. \ref{fig:representation} (a) - (c). Each of them can be transformed into the others. Specifically, we have:
\begin{itemize}
    \item \textit{Spherical representation} $\mathbf{I}(\omega, \phi)$, where $\omega$ from $-180^\circ$ to $180^\circ$ denotes the longitude and $\phi$ indicates the latitude from $-90^\circ$ to $90^\circ$ of a pixel. In practice, cos and sin transforms are adopted for $\omega$ and $\phi$, respectively, regarding the periodical property for traversal within an image. 
    \item \textit{Cubemap projection} treats $\mathbf{I}$ as a set of general 2D images, which are the faces of a cubic. In detail, we have $\mathbf{I}=\{\mathbf{i}_F, \mathbf{i}_L, \mathbf{i}_B, \mathbf{i}_R, \mathbf{i}_U, \mathbf{i}_{D}\}$, where each image $\mathbf{i} \in \mathbb{R}^{C \times H_{\mathbf{i}} \times W_{\mathbf{i}}}$ can be viewed as a general NFoV image, where $H_{\mathbf{i}}$ and $W_{\mathbf{i}}$ denote the height and the width of a face, respectively, and $C$ is the number of channels. 
    \item  \textit{Equirectangular projection} maps $\mathbf{I}$ to a general image in $\mathbb{R}^{C\times H_{\mathbf{I}}\times W_{\mathbf{I}}}$, where $H_{\mathbf{I}}$ and $W_{\mathbf{I}}$ indicate height and width, respectively.  Compared to cubemap projection, equirectangualr maps the entire spherical 360-degree image into a single rectangular grid, characterized by its noticeable pixel distortation around the top and bottom regions. 
\end{itemize}
As spherical representation inherently conforms to the 360-degree geometry coordinates, we project these geometry information to the cubemap form as shown in Fig. \ref{fig:representation} (d) - (e). 
We denote such geometry information as $\mathbf{\Gamma} = \{\mathbf{\gamma}_F, \mathbf{\gamma}_L, \mathbf{\gamma}_B, \mathbf{\gamma}_R, \mathbf{\gamma}_U, \mathbf{\gamma}_{D}\}$, where $\mathbf{\gamma}_{\boldsymbol{\cdot}} \in \mathbb{R}^{2 \times H_{\mathbf{i}} \times W_{\mathbf{i}}}$ contains the geometry information of a cubic face.

Given an NFoV image $\mathbf{X} \in \mathbb{R}^{C \times H_\mathbf{X} \times W_\mathbf{X}}$, such as a 2D image taken by smartphones, where $H_\mathbf{X}$ and $W_\mathbf{X}$ are its height and width, respectively; and a text guidance with its embedding $\mathbf{T} \in  \mathbb{R}^{C_{T} \times L}$, where $C_{T}$ is the dimension of textual feature and $L$ is the length of text guidance, our method aims to synthesize a 360-degree image $\hat{\mathbf{I}}$ by given $\mathbf{X}$ and $\mathbf{T}$.

\subsection{Autoregressive Omni-Traversal for Outpainting}

The autoregressive process outpaints the given NFoV image $\mathbf{X}$ progressively to a complete 360-degree image $\hat{\mathbf{I}}$ under the guidance of the text $\mathbf{T}$. Specifically, each step completes a local NFoV view, which is extracted from the incomplete 360-degree image with an unknown region that is neighbouring to a known region. 

\noindent\textbf{Local View Projection \& Backprojection.} 
To leverage a wide range of NFoV domain knowledge such as the weights from large-scale pretrained weights on NFoV image datasets, we correspondingly retrieve a local view from a 360-degree image $\mathbf{I}$ centred at the location $\omega$ and $\phi$ as a forward projection. 
In detail, we project the local view in $\mathbf{I}$ to an NFoV image $\mathbf{X}$ via a gnomonic projection, denoted as $\mathbf{X} = O(\mathbf{I}, \omega_{\mathbf{X}}, \phi_{\mathbf{X}})$, where $\omega_{\mathbf{X}}$ and $\phi_{\mathbf{X}}$ are the centroid longitude and latitude of the local view.
Similarly, we have a backprojection - an inverse gnomonic projection maps the pixels in $\mathbf{X}$ back to $\mathbf{I}$ partially within the scope of the corresponding NFoV view, denoted as $ \tilde{\mathbf{I}} = O^{-1}(\mathbf{X}, \omega_{\mathbf{X}}, \phi_{\mathbf{X}})$. 
Note that the pixel value out of the scope of $\mathbf{X}$ in $\tilde{\mathbf{I}}$ is defined as $-inf$. 
Furthermore, we define an operator $\oplus$ for two 360-degree inputs: $\mathbf{I}_\alpha$
and $\mathbf{I}_\beta$ as:
\begin{equation}
\mathbf{I}_\alpha \ \oplus \ \mathbf{I}_\beta (\omega, \phi)\ =\ \left\{
\begin{array}{lr}
\mathbf{I}_\alpha(\omega,\ \phi)\ \text{if}\ \mathbf{I}_\alpha(\omega,\ \phi)\ \neq\ -inf, \\
\mathbf{I}_\beta(\omega,\ \phi)\ \text{otherwise},
\end{array}
\right.
\end{equation}
which is used for attaching a newly generated partial 360-degree image to the current incomplete image. 

\noindent\textbf{Single-Step Outpainting}.
In a single-step outpainting, without loss of generality, for the $k^\text{th}$ step, an incomplete NFoV image $\bar{\mathbf{X}}_{k} = O(\bar{\mathbf{I}}_{k}, \omega_{\bar{\mathbf{X}}_{k}}, \phi_{\bar{\mathbf{X}}_{k}})$ is retrieved from an incomplete 360-degree image $\bar{\mathbf{I}}_{k}$.
Particularly, we denote a conditioned outpainting model $F_\mathbf{\Theta}$, where $\mathbf{\Theta}$ are learn-able parameters. $F_\mathbf{\Theta}$ estimates the unknown pixels in $\bar{\mathbf{X}}_{k}$, where the outpainted results is denoted as $\hat{\mathbf{X}}_{k} = F_\mathbf{\Theta}(\bar{\mathbf{X}}_{k}, \bar{\mathbf{I}}_{k}, \mathbf{T})$. 
The estimation $\hat{\mathbf{X}}_{k}$ is then backprojected to 360-degree view and a 360-degree outpainted estimation can be obtained as $\hat{\mathbf{I}}_{k} =  O^{-1}(\hat{\mathbf{X}}_{k},\omega_{\hat{\mathbf{X}}_{k}},\phi_{\hat{\mathbf{X}}_{k}}) \oplus \bar{\mathbf{I}}_{k}$. Note that $\omega_{\hat{\mathbf{X}}_{k}} = \omega_{\bar{\mathbf{X}}_{k}}$ and $\phi_{\hat{\mathbf{X}}_{k}} = \phi_{\bar{\mathbf{X}}_{k}}$ as $\hat{\mathbf{X}}_{k}$ retains its omni-geometry location. 
More details about $F_\mathbf{\Theta}$ can be found in the subsequent discussions.

Generally, $\bar{\mathbf{I}}_{1} = O^{-1}(\mathbf{X}, \omega_\mathbf{X}, \phi_\mathbf{X})$ is initialized with the input NFoV image $\mathbf{X}$. 
To optimize $F_\mathbf{\Theta}$, 
the known pixels in $\bar{\mathbf{X}}_{k}$ and $\bar{\mathbf{I}}_{k}$ can be extracted from the ground truth $\mathbf{I}$;  
we denote $\mathbf{X}_{k}$ and $\mathbf{I}_{k}$ as the ground truth
for the $k^\text{th}$ step. 
For inference, the known pixels in $\bar{\mathbf{X}}_{k}$ and $\bar{\mathbf{I}}_{k}$ can be based on the accumulated estimations $\hat{\mathbf{X}}_{k-1}$ and $\hat{\mathbf{I}}_{k-1}$, respectively. 

\noindent \textbf{Autoregressive Outpainting}.
Following an autoregressive stochastic process, a 360-degree image can be produced progressively:
\begin{equation}
\label{eqn:ar1}
    p(\mathbf{I}|\mathbf{T}) = \prod_k{p(\mathbf{I}_{k}|{\mathbf{I}}_{< k}, \mathbf{T})},         
\end{equation}
where ${\mathbf{I}}_{<k}$ indicates ${\mathbf{I}}_{1}$,..., ${\mathbf{I}}_{k-1}$, which are incomplete 360-degree images. 
As our proposed method mainly outpaints a small portion of unknown pixel in an incomplete 360-degree image, Eq. (\ref{eqn:ar1}) can be written with the Markov property: 
\begin{equation}
\label{eqn:ar2}
    p(\mathbf{I} |\mathbf{T}) = \prod_k{p({\mathbf{I}}_{k} |\bar{\mathbf{I}}_{k}, \mathbf{T} )} = \prod_k{p({\mathbf{X}}_{k} |\ \bar{\mathbf{I}}_{k}, \mathbf{T})}.
\end{equation}
In line with the single-step outpainting, $F_\mathbf{\Theta}$ is used to compute the conditioned probability terms as:
\begin{equation}
\label{eqn:ar3}
    p(\mathbf{I} |\mathbf{T}) \approx \prod_k{F_\mathbf{\Theta}(\bar{\mathbf{X}}_k, \bar{\mathbf{I}}_{k}, \mathbf{T})}.
\end{equation}
To this end, an autoregressive stochastic process has been formulated for 360-degree image generation. 
Note that we use an incremental pathway to identify $\bar{\mathbf{X}}_k$ that prioritizes the generation process along the longitude direction.

\subsection{Global-Local Conditioning by Omni-Aware Open-Vocabulary Guidance} 
AOG-Net incorporates multiple conditions to ensure its alignment to user text guidances and known NFoV views regarding the omni-geometry.  Specifically, in each autogressive step, a global-local conditioning mechanism is devised to thoroughly capture the following conditions:
\begin{itemize}
    \item Text guidance $\mathbf{c}_\text{text}$: a text encoder $\mathcal{E}_\text{text}$ encodes user text description $\textbf{T}$, which is based on the CLIP pre-trained textual model and enables an open-vocabulary paradigm to align the text features within a latent space shared with visual patterns. Note that this text guidance remains constant for each autoregressive step $k$ and acts as a global context. However, it can be modified according to user preferences to adjust during the generation process. 
    \item Omni-visual guidance $\mathbf{c}_{\text{360},k}$: a visual encoder $\mathcal{E}_\text{360}$, which leverages the CLIP pre-trained visual model, transforms a 360-degree image into the latent space that is shared with  $\mathbf{c}_\text{text}$. 
    Specifically, we encode each face in the cubemap representation of $\bar{\mathbf{I}}_k$ and denote the results as $\mathbf{c}_{\text{360},k} = \{\mathbf{c}_{F,k}, \mathbf{c}_{L,k}, \mathbf{c}_{B,k}, \mathbf{c}_{R,k}, \mathbf{c}_{U,k}, \mathbf{c}_{D,k}\}$. 
    \item NFoV guidance $\mathbf{c}_{\text{NFoV},k}$: a visual encoder $\mathcal{E}_\text{NFoV}$ encodes the incomplete NFoV image $\bar{\mathbf{X}}_k$ jointly with the 360-degree image $\bar{\mathbf{I}}_k$ in its cubemap form aiming for a omni-visual local latent representation. 
    \item Omni-geometry guidance $\mathbf{c}_{\text{geometry},k}$: an omni-geometry encoder $\mathcal{E}_\text{geometry}$ formulates the geometry  $\bar{\mathbf{\gamma}}_k$ of an incompleted local NFoV image $\bar{\mathbf{X}}_k$,  jointly with $\mathbf{\Gamma}$, to introduce the omni-geometry information for outpainting. 
\end{itemize}

\noindent \textbf{Global-Local Conditioning.} 
This module aligns the derived conditions for 360-degree outpainting through both a global and a local stream, leveraging cross-attention mechanisms. Globally, the incomplete 360-degree visual guidance $\mathbf{c}_{\text{360},k}$ is cross-referenced with the text guidance $\mathbf{c}_\text{text}$ to guarantee alignment between the content that already presents in $\mathbf{c}_{\text{360},k}$ and the content awaiting generation. 
Intuitively, we adopt a cross-attention based transformer for this purpose by treating the query as visual conditions $\mathbf{c}_{\text{360},k}$, while the value and key are as text conditions $\mathbf{c}_\text{text}$. We denote the results as a global condition $\mathbf{c}_{\text{global},k}$. 

Likewise, the local stream incorporates the NFoV guidance and the omni-geometry guidance using a transformer grounded in cross-attention. This integration facilitates the local fine-grained details during the generation process. 
Specifically, the query adopts the NFoV condition $\mathbf{c}_{\text{NFoV},k}$ supplemented by $\mathbf{c}_{\text{geometry},k}$, while the value and the key are with the 360-degree visual guidance $\mathbf{c}_{\text{360},k}$ supplemented by $\mathbf{c}_{\text{geometry},k}$. The resultant local condition is denoted as $\mathbf{c}_{\text{local},k}$. 

\subsection{Omni-Aware Diffusion for Outpainting}

Leveraging the recent success of the diffusion approach for NFoV content generation, in each autoregressive step, $F_\mathbf{\Theta}$ employs a stable diffusion backbone \cite{rombach2022highresolution}, incorporating the conditions $\mathbf{c}_{\text{global},k}$ and $\mathbf{c}_{\text{local},k}$. 
For the $k^\text{th}$ atuoregressive step, in a diffusion, we further denote $t$ as the diffusion temporal index and $\epsilon_\mathbf{\Theta}(\mathbf{z}_t, t)$ as the predicted noise introduced in the $t^\text{th}$ step, where $\epsilon_\mathbf{\Theta}$ is a U-Net. 
To optimize $\epsilon_\mathbf{\Theta}$, we minimize the following loss function:
\begin{equation}
    \mathcal{L}:=\mathbb{E}_{\epsilon_t \sim \mathcal{N}(0,1), t}\left[\left\|\epsilon_t-\epsilon_\theta\left(\mathbf{z}_t, t, \tau_\mathbf{\Theta}(\mathbf{c}_{\text{global},k},\mathbf{c}_{\text{local},k})\right)\right\|_2^2\right],
\end{equation}
where $\tau_{\mathbf{\Theta}}$ maps the conditions to guide the denoising process in the latent space via a cross-attention mechanism \cite{vaswani2023attention}.

\section{Experiments \& Discussions}

\subsection{Datasets} 
\textbf{360-Degree Images.} Following the existing studies \cite{akimoto2022diverse, somanath2021hdr}, we evaluate our proposed method with the LAVAL indoor HDR dataset \cite{gardner2017learning} for the 360-degree indoor image generation setting, which contains 2,233 360-degree images for extensive indoor scenes with a resolution of $7,768 \times 3,884$.  For a fair comparison, we used the official training and testing split in our experiments, in which we have 1,921 training samples and 312 testing samples.

For the outdoor setting, we utilize the LAVAL outdoor HDR dataset \cite{zhang2017learning}, which contains 210 360-degree images  with a resolution of $7,768 \times 3,884$.  In this setting, we randomly sample 170 images as the training split and 40 images for testing purpose.
In the training of both settings, the resolution of 360-degree image is downsampled to $4,096 \times 2048$ for computation efficiency.

\noindent \textbf{Text Captioning.} 
As the lack of text captions in both datasets, we adopted a large-scale captioning model BLIP2 \cite{li2023blip2} to generate captions for 360-degree images. 
We first caption an image in its equirectangular form to obtain an overall text guidance with an average of 5-6 words. Next, we caption the horizontal faces individually of its cubemap to obtain additional text guidances.

\noindent \textbf{Data Augmentation.} To increase the diversity of the 360-degree images generated, we augmented the training 360-degree image samples by adopting random clockwise rotation based on the intrinsic horizontal cyclicity. To improve the diversity of text guidance, besides randomly swapping words with TextAttack \cite{morris2020textattack}, we randomly combines the overall text guidance and one randomly selected text guidance associated with a face of the cubemap during training.

\subsection{Implementation Details}

\noindent \textbf{Pre-Trained Models \& Network Architecture.} In our experiment, we adopted the pretrained Stable Diffusion generative prior for each autoregressive generation step. In addition, We utilized the visual encoder and the text encoder of OpenCLIP \cite{cherti2022reproducible} for $\mathcal{E}_\text{360}$ and $\mathcal{E}_\text{text}$, respectively. 
We utilized T2I-Adapter \cite{mou2023t2iadapter} as the architecture for NFoV guidance encoder $\mathcal{E}_\text{NFoV}$ and omni-geometry guidance encoder $\mathcal{E}_\text{geometry}$. 
In both local stream and global stream, we utilized a 16-layer cross-attion base transformer to compute $\mathbf{c}_{\text{local},k}$ and $\mathbf{c}_{\text{global},k}$ respectively.

\noindent \textbf{Training and Inference.} AOG-Net was trained using an AdamW optimizer \cite{loshchilov2019decoupled} with $\beta_1 = 0.9$ and $\beta_2 = 0.999$. It was trained for 240 epochs, with learning rate $1 \times 10^{-4}$ and batch size 1. 
For inference, we leveraged DPM-Solver++ \cite{lu2023dpmsolver} as sampler with a step set to 25 and classifier-free-guidance \cite{ho2022classifierfree} scale set to 2.5. 
All experiments were conducted on an Nvidia RTX 3090.

\begin{figure*}[h]
\centering
\includegraphics[width=0.99\textwidth]{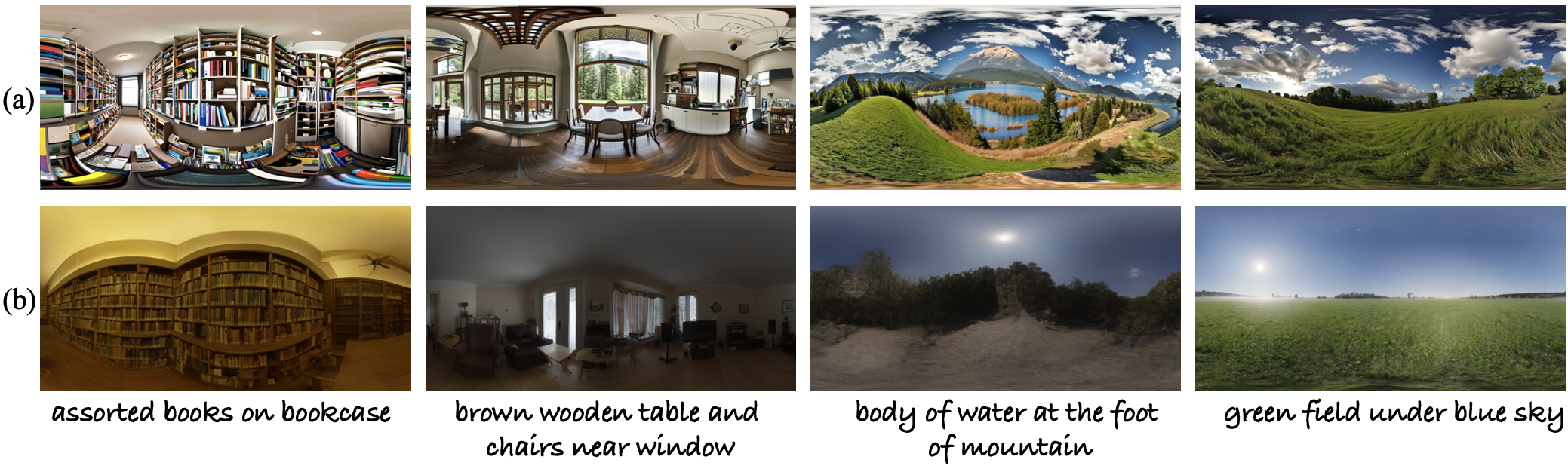} 
\caption{Comparison between Text2Light for indoor and outdoor settings. (a) Ours. (b) Text2Light.}
\label{fig:compare_text2light}
\end{figure*}

\subsection{Comparison with State of the Art}

\noindent \textbf{Baselines.} Our method is compared with the recent state-of-the-art 360-degree image outpainting methods from three perspectives. 1) NFoV image guided generation methods without text guidance: ImmerseGAN \cite{Dastjerdi_2022}, OmniDreamer\cite{akimoto2022diverse}  and EnvMapNet \cite{somanath2021hdr}. For a fair comparison, the text guidance in our method is set as a blank prompt. 
2) Text-guided generation method without NFoV guidance - Text2Light \cite{chen2023text2light}. 
In this case, we generated our initial input NFoV image using Stable Diffusion Outpainting model for our method. 
3) NFoV image and text guided generation method \cite{Dastjerdi_2022}. 

\noindent \textbf{Evalution Metrics.} To quantitatively evaluate our AOG-Net, we adopted LPIPS \cite{zhang2018unreasonable} and  Fréchet Inception Distance (FID) \cite{heusel2018gans} as the evaluation metrics to measure the similarity of latent representations between the generated 360-degree images and the ground truth. 
To evaluate the semantic consistency (SC) between the generated 360-degree image and the input text guidance, we compared the similarity between input text guidance and the captioning texts obtained from the generated image. Specifically, we leveraged a large-scale captioning model BLIP2 \cite{li2023blip2} and computed the similarity with sentence embeddings \cite{reimers-2019-sentence-bert} for this purpose.
In addition, we leverage Inception Score (IS) \cite{salimans2016improved} to measure the quality of the generated images as Text2Light does not involve ground truth images.

\begin{table}[ht]
\begin{center}
\begin{tabularx}{\columnwidth}{l@{\extracolsep{\fill}}cc} 
 \hline
 Method &  FID $\downarrow$ & LPIPS$\downarrow$\\ 
 \hline
 \hline
 \text{Indoor setting} &  & \\ 
 \hline
 \hline
 SIG-SS (\citeyear{gardner2017learning}) &  197.4 & -\\ 
 EnvMapNet (\citeyear{somanath2021hdr}) &  52.70& -\\
 OmniDreamer (\citeyear{akimoto2022diverse}) &  46.15& \underline{0.45}\\
 ImmenseGAN(\citeyear{Dastjerdi_2022}) &  \underline{42.78} & -\\
 AOG-Net (Ours) &  \textbf{38.60} & \textbf{0.37}\\ 
 \hline
 \hline
 \text{Outdoor setting} &  & \\ 
 \hline
 \hline
 OmniDreamer (\citeyear{akimoto2022diverse}) & \underline{24.5} & \underline{0.41} \\
 AOG-Net (Ours) &  \textbf{18.4}  & \textbf{0.36}  \\ 
 \hline
\end{tabularx}
\end{center}
\caption{Comparison with the state-of-the-art methods using NFoV image guidance. }
\label{tab:quant2}
\end{table}

\begin{table}[ht]
\begin{center}
\begin{tabularx}{\columnwidth}{l@{\extracolsep{\fill}}cc}
 \hline
 Method & SC $\uparrow$ & IS$\uparrow$ \\ 
 \hline
 \hline
 \text{Outdoor setting} &  & \\ 
 \hline
 \hline
 Text2Light (\citeyear{chen2023text2light}) & \underline{0.45} &  \underline{3.9}\\
 AOG-Net (Ours)&  \textbf{0.53}  & \textbf{4.2}  \\ 
\hline
\hline
\text{Indoor setting} &  & \\ 
\hline
\hline
 Text2Light (\citeyear{chen2023text2light}) & \underline{0.33} &  \underline{4.5}\\
 AOG-Net (Ours) &  \textbf{0.36}  & \textbf{5.1}  \\ 
 \hline
\end{tabularx}
\end{center}
\caption{Comparison with the state-of-the-art methods using text guidance.}
\label{tab:quant3}
\end{table}

\begin{figure}[ht]
\centering
\includegraphics[width=1.\columnwidth]{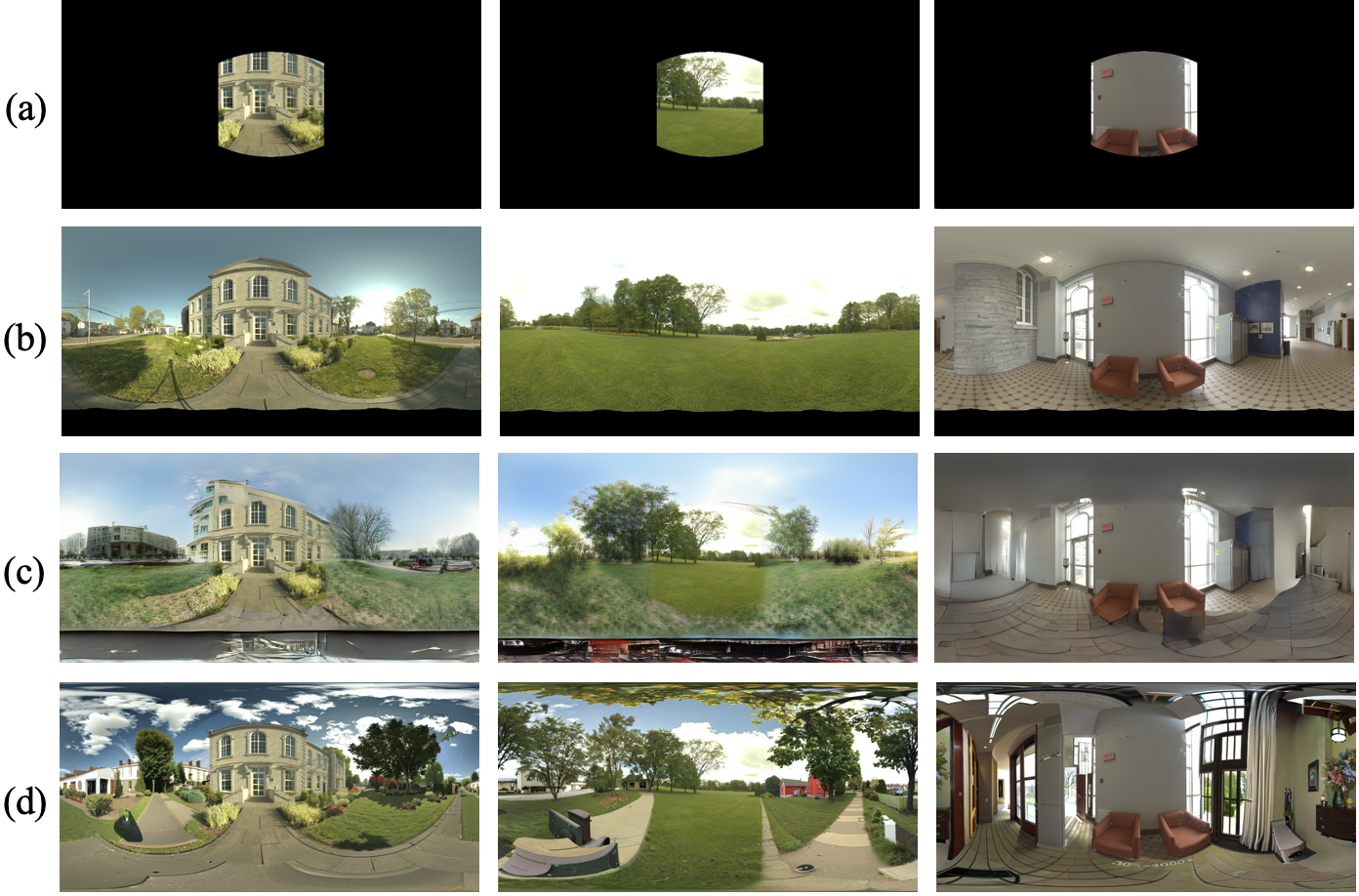} 
\caption{Qualitative examples with input and ground truth. (a) Input, $90^{\circ}$ in both longitude and latitude direction. (b) grount truth. (c) OmniDreamer. (d) AOG-Net (Ours).}
\label{fig:qualitative_omni}
\end{figure}

\noindent \textbf{Overall Performance.}
For the methods requiring an initial NFoV image as guidance, 
our method achieve the best performance as shown in Table \ref{tab:quant2}. It achieves an FID score 35.6 and an LPIPD value 0.37 under an indoor setting and an FID score 18.4 and an LPIPS value 0.36 under an outdoor setting. Note that only OmniDreamer conducted evaluation for outdoor setting in the literature. 
As shown in the first example (in the first column) in Fig. \ref{fig:qualitative_omni}, 
our method outpaints the house and the garden smoothly, while OmniDreamer make the neighbouring region of the house smudged with sudden color changes in the garden. For the third example  (in the third column), our method is able to deliver detailed outpainting regarding objects compared to Omnidremaer. 

Regarding the comparison with text-conditioned method - Text2Light with open vocabulary text guidances, the performance metrics are listed in Table \ref{tab:quant3}. Our method outperform Text2Light with an SC score 0.53 and an IS value 4.2 for an outdoor setting and an SC score 0.36 and an IS value 5.1 for an indoor setting. 
Due to the complexity of the indoor setting and the lack of in-depth text description, the semantic consistence of both methods drop. However, our method still provide overall better semantically consistent images with higher image quality.
As depicted in Figure \ref{fig:compare_text2light}, our method is able to produce visually appealing images, while the images of Text2Light \cite{chen2023text2light} are much dimmer, leading to degenerated visual quality and lack of details. Additionally, under the outdoor setting, our method generates 360-degree image with fine-grained details (such as trees, grasses), while the Text2Light produces smudged-out patterns in images (in the third column of Fig. \ref{fig:compare_text2light}).

For ImmenseGAN, which leverages both NFoV and text guidances, our method peforms superior under the indoor setting according to the metrics the authors reported in their work. ImmenseGAN was trained with a private large-scale dataset and the authors did not evaluate their method under an outdoor setting. Note that our method leverages the pretrained diffusion models and only requires 40 randomly selected training samples to achieve its current performance.

\subsection{Ablation Study}

Ablation studies are conducted to demonstrate the effectiveness of individual components in AOG-Net. 
The results are listed in Table \ref{tab:ablation} and an example is shown in Fig. 6.

\begin{figure}[ht]
\centering
\includegraphics[width=1.\columnwidth]{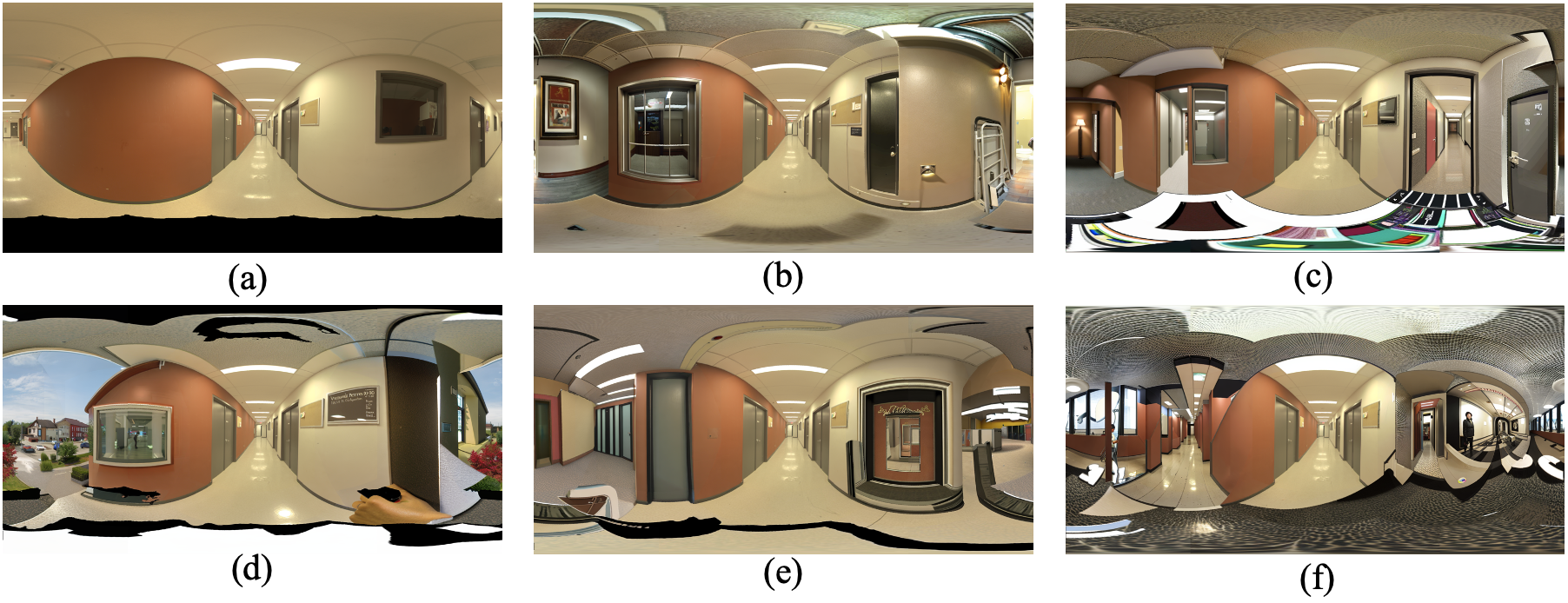} 
\caption{Ablation study. (a) Ground truth. (b) AOG-Net. (c) w/o global condition. (d) w/o local condition. (e) w/o geometry condition. (f) Autoregressive w/ backbone only.}
\label{fig:ablation}
\end{figure}

\noindent \textbf{Global Guidance.} The global guidance $\mathbf{c}_{global}$ and its related components are removed from the pipeline. The backbone model is only guided by CLIP text guidance embedding and $\mathbf{c}_{local}$. Although the presence of the text guidance provides a coarse global condition, the overall semantic consistency is decreased. As shown in Fig.\ref{fig:ablation} (c), the model fails to deliver a consistent floor texture without global guidance.

\noindent \textbf{Local Guidance.} 
The local guidance $\mathbf{c}_{local}$ and its related components are excluded from our method.
While the semantic consistency between the outputs and the text prompts is only slightly affected, the deterioration in the quality of the outpainted images is significant. In Fig. \ref{fig:ablation} (d), there are various artifacts such as black patches and human hands.

\noindent \textbf{Geometry Guidance.} In this setting, we removes all 360-degree geometry information  $\mathbf{c}_\text{geo}$ in computing $\mathbf{c}_{local}$, which would only rely on pixel-wise semantics to connect distant patches. 
The results reveal minimal effects on SC, but there is a notable decrease in image quality. As illustrated in Fig. \ref{fig:ablation} (e), black patterns appear on the floor and the ceiling's color lacks consistency with distant regions.

\noindent \textbf{Backbone Only.} 
In this setting, only the pre-trained Stable Diffusion backbone is employed in a traditional manner, integrating the NFoV input image and text guidance. This settings produces the poorest SC values and FID scores, suggesting that the outpainted 360-degree images are of low quality and misaligned with the text guidance. Referring to the generation example shown in Fig. \ref{fig:ablation} (f), the model struggles to produce a text-consistent and sharp 360-degree image outpainting, with evident localized artifacts.

\begin{table}[ht]
\begin{center}
\begin{tabularx}{\columnwidth}{l@{\extracolsep{\fill}}ccc}
 \hline
 Method & LPIPS$\downarrow$ & FID$\downarrow$ & SC$\uparrow$\\ 
 \hline
\hline
 AOG-Net (Ours) &  \underline{0.37} & \textbf{35.6} & \textbf{0.72}\\ 
 w/o global condition & 0.38  & 40.08  & 0.70\\
  w/o local condition & 0.40  &  47.46 & 0.71 \\
 w/o geometry condition & \textbf{0.36}   &  \underline{37.2} & \textbf{0.72}\\ 
 Autoregressive w/ backbone only  &  0.43  & 67.4 & 0.61 \\ 
 \hline
\end{tabularx}
\end{center}
\caption{Ablation study on the Laval Indoor HDR dataset.
}
\label{tab:ablation}
\end{table}

\begin{figure}[ht]
\centering
\includegraphics[width=1.\columnwidth]{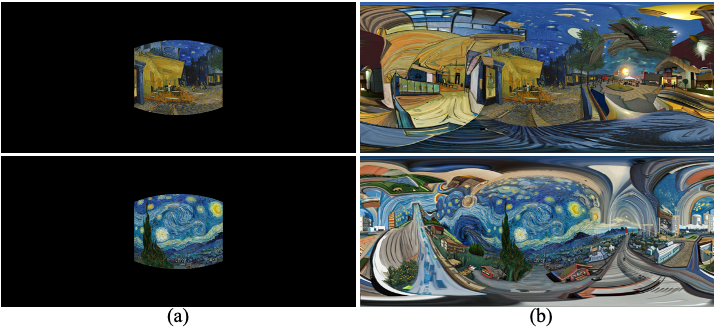} 
\caption{Open-image conditioned generation results, with the prompt ``a 360 image of a city, oil painting , ultracolorful, impressionist style, Van Gogh style". (a) Input. (b) Output.}
\label{fig:van}
\end{figure}
\subsection{Generalization} We further explore an open-image conditioned task. Our method is required to outpaint unseen oil painting artworks to 360-degree images with text guidances. As shown in Fig. \ref{fig:van}, the generated images are consistent with the style of input NFoV artworks, demonstrating the potential of our method in accepting out-of-domain NFoV image as conditions.

\subsection{Limitation and Future Work}
AOG-Net relies on a pre-trained backbone model, which introduces two primary limitations.
Firstly, AOG-Net is somewhat constrained by the data on which the backbone model was pre-trained, potentially, this method may suffer from the internal biases introduced by the backbone diffusion model.   
Secondly, the diffusion model's prolonged inference time affects its utility in applications that require real-time performance.
Future endeavors might focus on developing a backbone or the exploration of conditional 360-degree image generation similar to \cite{stan2023ldm3d}.
Finally, by capitalizing on the autoregressive characteristics, our method has the potential to be extended to facilitate text-guided 360-degree video generation.

\section{Conclusion}
In this work, we present a novel deep learning method AOG-Net for 360-degree image generation with an autoregressive scheme guided by  NFoV images and open vocabulary text prompts. 
A global-local conditioning mechanism is devised to adaptively encode guidances considering omni-directional properties. With these design, AOG-Net is able to generate realistic 360-degree image with fine details while aligning with the text guidance.
Comprehensive experiments demonstrate the effectiveness of AOG-Net. 

\section*{Acknowledgments}
This study was partially supported by Australian Research Council (ARC) grant \#DP210102674.

\bibliography{aaai24}

\begin{thebibliography}{40}
\providecommand{\natexlab}[1]{#1}

\bibitem[{Akimoto et~al.(2019)Akimoto, Kasai, Hayashi, and Aoki}]{360ic}
Akimoto, N.; Kasai, S.; Hayashi, M.; and Aoki, Y. 2019.
\newblock 360-Degree Image Completion by Two-Stage Conditional Gans.
\newblock In \emph{IEEE International Conference on Image Processing}.

\bibitem[{Akimoto, Matsuo, and Aoki(2022)}]{akimoto2022diverse}
Akimoto, N.; Matsuo, Y.; and Aoki, Y. 2022.
\newblock Diverse Plausible 360-Degree Image Outpainting for Efficient 3DCG
  Background Creation.
\newblock In \emph{IEEE/CVF Conference on Computer Vision and Pattern
  Recognition}.

\bibitem[{Chang et~al.(2022)Chang, Zhang, Jiang, Liu, and
  Freeman}]{chang2022maskgit}
Chang, H.; Zhang, H.; Jiang, L.; Liu, C.; and Freeman, W.~T. 2022.
\newblock MaskGIT: Masked Generative Image Transformer.
\newblock In \emph{IEEE Conference on Computer Vision and Pattern Recognition}.

\bibitem[{Chen et~al.(2020)Chen, Radford, Child, Wu, Jun, Luan, and
  Sutskever}]{imaggpt}
Chen, M.; Radford, A.; Child, R.; Wu, J.; Jun, H.; Luan, D.; and Sutskever, I.
  2020.
\newblock Generative Pretraining from Pixels.
\newblock In \emph{International Conference on Machine Learning}.

\bibitem[{Chen, Wang, and Liu(2022)}]{chen2023text2light}
Chen, Z.; Wang, G.; and Liu, Z. 2022.
\newblock Text2Light: Zero-Shot Text-Driven HDR Panorama Generation.
\newblock \emph{ACM Transactions on Graphics}.

\bibitem[{Cherti et~al.(2023)Cherti, Beaumont, Wightman, Wortsman, Ilharco,
  Gordon, Schuhmann, Schmidt, and Jitsev}]{cherti2022reproducible}
Cherti, M.; Beaumont, R.; Wightman, R.; Wortsman, M.; Ilharco, G.; Gordon, C.;
  Schuhmann, C.; Schmidt, L.; and Jitsev, J. 2023.
\newblock Reproducible scaling laws for contrastive language-image learning.
\newblock In \emph{IEEE/CVF Conference on Computer Vision and Pattern
  Recognition}.

\bibitem[{Dastjerdi et~al.(2022)Dastjerdi, Hold-Geoffroy, Eisenmann,
  Khodadadeh, and Lalonde}]{Dastjerdi_2022}
Dastjerdi, M. R.~K.; Hold-Geoffroy, Y.; Eisenmann, J.; Khodadadeh, S.; and
  Lalonde, J.-F. 2022.
\newblock Guided Co-Modulated {GAN} for 360{\textdegree} Field of View
  Extrapolation.
\newblock In \emph{International Conference on 3D Vision}.

\bibitem[{Dhariwal and Nichol(2021)}]{dhariwal2021diffusion}
Dhariwal, P.; and Nichol, A. 2021.
\newblock Diffusion models beat gans on image synthesis.
\newblock \emph{Advances in Neural Information Processing Systems}.

\bibitem[{Esser, Rombach, and Ommer(2021)}]{esser2020taming}
Esser, P.; Rombach, R.; and Ommer, B. 2021.
\newblock Taming transformers for high-resolution image synthesis.
\newblock In \emph{IEEE/CVF Conference on Computer Vision and Pattern
  Recognition}.

\bibitem[{Gardias, Arthur, and Sun(2020)}]{gardias2020enhanced}
Gardias, P.; Arthur, E.; and Sun, H. 2020.
\newblock Enhanced Residual Networks for Context-based Image Outpainting.
\newblock arXiv:2005.06723.

\bibitem[{Gardner et~al.(2017)Gardner, Sunkavalli, Yumer, Shen, Gambaretto,
  Gagné, and Lalonde}]{gardner2017learning}
Gardner, M.-A.; Sunkavalli, K.; Yumer, E.; Shen, X.; Gambaretto, E.; Gagné,
  C.; and Lalonde, J.-F. 2017.
\newblock Learning to Predict Indoor Illumination from a Single Image.
\newblock arXiv:1704.00090.

\bibitem[{Hara, Mukuta, and Harada(2021)}]{hara2020spherical}
Hara, T.; Mukuta, Y.; and Harada, T. 2021.
\newblock Spherical Image Generation from a Single Image by Considering Scene
  Symmetry.
\newblock \emph{AAAI Conference on Artificial Intelligence}.

\bibitem[{Heusel et~al.(2017)Heusel, Ramsauer, Unterthiner, Nessler, and
  Hochreiter}]{heusel2018gans}
Heusel, M.; Ramsauer, H.; Unterthiner, T.; Nessler, B.; and Hochreiter, S.
  2017.
\newblock GANs Trained by a Two Time-Scale Update Rule Converge to a Local Nash
  Equilibrium.
\newblock In Guyon, I.; Luxburg, U.~V.; Bengio, S.; Wallach, H.; Fergus, R.;
  Vishwanathan, S.; and Garnett, R., eds., \emph{2017 Advances in Neural
  Information Processing Systems}.

\bibitem[{Ho, Jain, and Abbeel(2020)}]{ho2020denoising}
Ho, J.; Jain, A.; and Abbeel, P. 2020.
\newblock Denoising diffusion probabilistic models.
\newblock \emph{Advances in Neural Information Processing Systems}.

\bibitem[{Ho and Salimans(2022)}]{ho2022classifierfree}
Ho, J.; and Salimans, T. 2022.
\newblock Classifier-Free Diffusion Guidance.
\newblock arXiv:2207.12598.

\bibitem[{Hoorick(2020)}]{vanhoorick2020image}
Hoorick, B.~V. 2020.
\newblock Image Outpainting and Harmonization using Generative Adversarial
  Networks.
\newblock arXiv:1912.10960.

\bibitem[{Isola et~al.(2017)Isola, Zhu, Zhou, and
  Efros}]{isola2018imagetoimage}
Isola, P.; Zhu, J.-Y.; Zhou, T.; and Efros, A.~A. 2017.
\newblock Image-to-image translation with conditional adversarial networks.
\newblock In \emph{IEEE Conference on Computer Vision and Pattern Recognition}.

\bibitem[{Kang et~al.(2023)Kang, Zhu, Zhang, Park, Shechtman, Paris, and
  Park}]{kang2023scaling}
Kang, M.; Zhu, J.-Y.; Zhang, R.; Park, J.; Shechtman, E.; Paris, S.; and Park,
  T. 2023.
\newblock Scaling up gans for text-to-image synthesis.
\newblock In \emph{IEEE/CVF Conference on Computer Vision and Pattern
  Recognition}.

\bibitem[{Li et~al.(2023)Li, Li, Savarese, and Hoi}]{li2023blip2}
Li, J.; Li, D.; Savarese, S.; and Hoi, S. 2023.
\newblock Blip-2: Bootstrapping language-image pre-training with frozen image
  encoders and large language models.
\newblock \emph{arXiv preprint arXiv:2301.12597}.

\bibitem[{Loshchilov and Hutter(2019)}]{loshchilov2019decoupled}
Loshchilov, I.; and Hutter, F. 2019.
\newblock Decoupled Weight Decay Regularization.
\newblock In \emph{International Conference on Learning Representations}.

\bibitem[{Lu et~al.(2023)Lu, Zhou, Bao, Chen, Li, and Zhu}]{lu2023dpmsolver}
Lu, C.; Zhou, Y.; Bao, F.; Chen, J.; Li, C.; and Zhu, J. 2023.
\newblock DPM-Solver++: Fast Solver for Guided Sampling of Diffusion
  Probabilistic Models.
\newblock arXiv:2211.01095.

\bibitem[{Lu, Chang, and Chiu(2021)}]{lu2021bridging}
Lu, C.-N.; Chang, Y.-C.; and Chiu, W.-C. 2021.
\newblock Bridging the visual gap: Wide-range image blending.
\newblock In \emph{IEEE/CVF Conference on Computer Vision and Pattern
  Recognition}.

\bibitem[{Mirza and Osindero(2014)}]{mirza2014conditional}
Mirza, M.; and Osindero, S. 2014.
\newblock Conditional Generative Adversarial Nets.
\newblock arXiv:1411.1784.

\bibitem[{Morris et~al.(2020)Morris, Lifland, Yoo, Grigsby, Jin, and
  Qi}]{morris2020textattack}
Morris, J.; Lifland, E.; Yoo, J.~Y.; Grigsby, J.; Jin, D.; and Qi, Y. 2020.
\newblock TextAttack: A Framework for Adversarial Attacks, Data Augmentation,
  and Adversarial Training in NLP.
\newblock In \emph{Conference on Empirical Methods in Natural Language
  Processing: System Demonstrations}.

\bibitem[{Mou et~al.(2023)Mou, Wang, Xie, Wu, Zhang, Qi, Shan, and
  Qie}]{mou2023t2iadapter}
Mou, C.; Wang, X.; Xie, L.; Wu, Y.; Zhang, J.; Qi, Z.; Shan, Y.; and Qie, X.
  2023.
\newblock T2I-Adapter: Learning Adapters to Dig out More Controllable Ability
  for Text-to-Image Diffusion Models.
\newblock arXiv:2302.08453.

\bibitem[{Radford et~al.(2021)Radford, Kim, Hallacy, Ramesh, Goh, Agarwal,
  Sastry, Askell, Mishkin, Clark et~al.}]{radford2021learning}
Radford, A.; Kim, J.~W.; Hallacy, C.; Ramesh, A.; Goh, G.; Agarwal, S.; Sastry,
  G.; Askell, A.; Mishkin, P.; Clark, J.; et~al. 2021.
\newblock Learning transferable visual models from natural language
  supervision.
\newblock In \emph{International Conference on Machine Learning}.

\bibitem[{Reimers and Gurevych(2019)}]{reimers-2019-sentence-bert}
Reimers, N.; and Gurevych, I. 2019.
\newblock Sentence-BERT: Sentence Embeddings using Siamese BERT-Networks.
\newblock In \emph{Conference on Empirical Methods in Natural Language
  Processing}.

\bibitem[{Rombach et~al.(2022)Rombach, Blattmann, Lorenz, Esser, and
  Ommer}]{rombach2022highresolution}
Rombach, R.; Blattmann, A.; Lorenz, D.; Esser, P.; and Ommer, B. 2022.
\newblock High-resolution image synthesis with latent diffusion models.
\newblock In \emph{IEEE/CVF Conference on Computer Vision and Pattern
  Recognition}.

\bibitem[{Sabini and Rusak(2018)}]{sabini2018painting}
Sabini, M.; and Rusak, G. 2018.
\newblock Painting Outside the Box: Image Outpainting with GANs.
\newblock arXiv:1808.08483.

\bibitem[{Salimans et~al.(2016)Salimans, Goodfellow, Zaremba, Cheung, Radford,
  and Chen}]{salimans2016improved}
Salimans, T.; Goodfellow, I.; Zaremba, W.; Cheung, V.; Radford, A.; and Chen,
  X. 2016.
\newblock Improved techniques for training gans.
\newblock \emph{Advances in Neural Information Processing Systems}.

\bibitem[{Sengupta et~al.(2019)Sengupta, Gu, Kim, Liu, Jacobs, and
  Kautz}]{sengupta2019neural}
Sengupta, S.; Gu, J.; Kim, K.; Liu, G.; Jacobs, D.~W.; and Kautz, J. 2019.
\newblock Neural inverse rendering of an indoor scene from a single image.
\newblock In \emph{IEEE/CVF International Conference on Computer Vision}.

\bibitem[{Somanath and Kurz(2021)}]{somanath2021hdr}
Somanath, G.; and Kurz, D. 2021.
\newblock HDR environment map estimation for real-time augmented reality.
\newblock In \emph{IEEE/CVF Conference on Computer Vision and Pattern
  Recognition}.

\bibitem[{Song, Meng, and Ermon(2021)}]{song2022denoising}
Song, J.; Meng, C.; and Ermon, S. 2021.
\newblock Denoising Diffusion Implicit Models.
\newblock In \emph{International Conference on Learning Representations}.

\bibitem[{Stan et~al.(2023)Stan, Wofk, Fox, Redden, Saxton, Yu, Aflalo, Tseng,
  Nonato, Muller et~al.}]{stan2023ldm3d}
Stan, G. B.~M.; Wofk, D.; Fox, S.; Redden, A.; Saxton, W.; Yu, J.; Aflalo, E.;
  Tseng, S.-Y.; Nonato, F.; Muller, M.; et~al. 2023.
\newblock LDM3D: Latent Diffusion Model for 3D.
\newblock \emph{arXiv preprint arXiv:2305.10853}.

\bibitem[{Vaswani et~al.(2017)Vaswani, Shazeer, Parmar, Uszkoreit, Jones,
  Gomez, Kaiser, and Polosukhin}]{vaswani2023attention}
Vaswani, A.; Shazeer, N.; Parmar, N.; Uszkoreit, J.; Jones, L.; Gomez, A.~N.;
  Kaiser, L.~u.; and Polosukhin, I. 2017.
\newblock Attention is All you Need.
\newblock In Guyon, I.; Luxburg, U.~V.; Bengio, S.; Wallach, H.; Fergus, R.;
  Vishwanathan, S.; and Garnett, R., eds., \emph{Advances in Neural Information
  Processing Systems}.

\bibitem[{Wang et~al.(2019)Wang, Tao, Shen, and Jia}]{wang2019srn}
Wang, Y.; Tao, X.; Shen, X.; and Jia, J. 2019.
\newblock Wide-Context Semantic Image Extrapolation.
\newblock In \emph{IEEE Conference on Computer Vision and Pattern Recognition}.

\bibitem[{Yao et~al.(2022)Yao, Gao, Yang, Sun, Zhang, and Huang}]{yao2022qotr}
Yao, K.; Gao, P.; Yang, X.; Sun, J.; Zhang, R.; and Huang, K. 2022.
\newblock Outpainting by Queries.
\newblock In \emph{European Conference on Computer Vision}.

\bibitem[{Yariv et~al.(2023)Yariv, Gat, Wolf, Adi, and
  Schwartz}]{yariv2023audiotoken}
Yariv, G.; Gat, I.; Wolf, L.; Adi, Y.; and Schwartz, I. 2023.
\newblock AudioToken: Adaptation of Text-Conditioned Diffusion Models for
  Audio-to-Image Generation.
\newblock arXiv:2305.13050.

\bibitem[{Zhang and Lalonde(2017)}]{zhang2017learning}
Zhang, J.; and Lalonde, J.-F. 2017.
\newblock Learning High Dynamic Range from Outdoor Panoramas.
\newblock In \emph{2017 IEEE International Conference on Computer Vision}.

\bibitem[{Zhang et~al.(2018)Zhang, Isola, Efros, Shechtman, and
  Wang}]{zhang2018unreasonable}
Zhang, R.; Isola, P.; Efros, A.~A.; Shechtman, E.; and Wang, O. 2018.
\newblock The unreasonable effectiveness of deep features as a perceptual
  metric.
\newblock In \emph{IEEE Conference on Computer Cision and Pattern Recognition}.

\end{thebibliography}

\end{document}